\documentclass[11pt,a4paper]{article}
\usepackage{emnlp2018}
\usepackage{times}
\usepackage{latexsym}
\usepackage{url}
\usepackage{enumerate}
\usepackage{graphicx}
\usepackage{booktabs}
\usepackage{colortbl}
\usepackage{subcaption}

\usepackage{tablefootnote}
\usepackage{longtable}

\usepackage{pgfplotstable} 
\usepackage{pgfplots}
\pgfplotsset{compat=1.14}

\usepackage[normalem]{ulem} 

\aclfinalcopy 

\newcommand{\grow}[1]{\rowcolor{gray!10} #1}

\title{Understanding Back-Translation at Scale}

\author{%
Sergey Edunov$^\bigtriangleup$
\quad Myle Ott$^\bigtriangleup$
\quad Michael Auli$^\bigtriangleup$
\quad David Grangier\hspace{1pt}$^\bigtriangledown$$^{*}$
\\
$^\bigtriangleup$Facebook AI Research, Menlo Park, CA \& New York, NY. \quad \\
$^\bigtriangledown$Google Brain, Mountain View, CA.
}
\date{}

\begin{document}
\maketitle

{\let\thefootnote\relax\footnotetext{*Work done while at Facebook AI Research.}}

\begin{abstract}
An effective method to improve neural machine translation with monolingual data is to augment the parallel training corpus with back-translations of target language sentences.
This work broadens the understanding of back-translation and investigates a number of methods to generate synthetic source sentences.
We find that in all but resource poor settings back-translations obtained via sampling or noised beam outputs are most effective.
Our analysis shows that sampling or noisy synthetic data gives a much stronger training signal than data generated by beam or greedy search.
We also compare how synthetic data compares to genuine bitext and study various domain effects.
Finally, we scale to hundreds of millions of monolingual sentences and achieve a new state of the art of 35 BLEU on the WMT'14 English-German test set.
\end{abstract}

\section{Introduction}

Machine translation relies on the statistics of large parallel corpora, i.e. datasets of paired sentences in both the source and target language.
However, bitext is limited and there is a much larger amount of monolingual data available. 
Monolingual data has been traditionally used to train language models which improved the fluency of statistical machine translation \citep{koehn:book}.

In the context of neural machine translation (NMT; \citealt{bahdanau:attention:2015}; \citealt{gehring:convs2s:2017}; \citealt{vaswani:transformer:2017}), there has been extensive work to improve models with monolingual data, including language model fusion~\cite{gulcehre:lm:2015,gulcehre:lm:2017}, back-translation~\cite{sennrich:backtranslation:2016} and dual learning~\cite{cheng:semi:2016,he:dual:2016}. 
These methods have different advantages and can be combined to reach high accuracy~\cite{hassan:parity:2018}.

We focus on back-translation (BT) which operates in a semi-supervised setup where both bilingual and monolingual data in the target language are available.
Back-translation first trains an intermediate system on the parallel data which is used to translate the target monolingual data into the source language. 
The result is a parallel corpus where the source side is \emph{synthetic} machine translation output while the target is genuine text written by humans.
The synthetic parallel corpus is then simply added to the real bitext in order to train a final system that will translate from the source to the target language.
Although simple, this method has been shown to be helpful for phrase-based translation~\citep{bojar:bt_pbmt:2011}, NMT~\citep{sennrich:backtranslation:2016,poncelas:investigatingBT:2018} as well as unsupervised MT~\citep{lample:unsupervised:2018}. 

In this paper, we investigate back-translation for neural machine translation at a large scale by adding hundreds of millions of back-translated sentences to the bitext.
Our experiments are based on strong baseline models trained on the public bitext of the WMT competition. 
We extend previous analysis \citep{sennrich:backtranslation:2016,poncelas:investigatingBT:2018} of back-translation in several ways. 
We provide a comprehensive analysis of different methods to generate synthetic source sentences and we show that this choice matters: 
sampling from the model distribution or noising beam outputs outperforms pure beam search, which is typically used, by 1.7 BLEU on average across several test sets.
Our analysis shows that synthetic data based on sampling and noised beam search provides a stronger training signal than synthetic data based on argmax inference.
We also study how adding synthetic data compares to adding real bitext in a controlled setup with the surprising finding that synthetic data can sometimes match the accuracy of real bitext.
Our best setup achieves 35 BLEU on the WMT'14 English-German test set by relying only on public WMT bitext as well as 226M monolingual sentences.
This outperforms the system of DeepL by 1.7 BLEU who train on large amounts of high quality non-benchmark data.
On WMT'14 English-French we achieve 45.6 BLEU.

\section{Related work}
\label{sec:related}

This section describes prior work in machine translation with neural networks as well as semi-supervised machine translation. 

\subsection{Neural machine translation}

We build upon recent work on neural machine translation which is typically a neural network with an encoder/decoder architecture. The encoder infers a continuous space representation of the source sentence, while the decoder is a neural language model conditioned on the encoder output. The parameters of both models are learned jointly to maximize the likelihood of the target sentences given the corresponding source sentences from a parallel corpus~\citep{sutskever:seq2seq:2014,cho:seq2seq:2014}. At inference, a target sentence is generated by left-to-right decoding.

Different neural architectures have been proposed with the goal of improving efficiency and/or effectiveness. This includes recurrent networks~\citep{sutskever:seq2seq:2014,bahdanau:attention:2015,luong:nmt:2015}, convolutional networks~\citep{kalchbrenner:bytenet:2016,gehring:convs2s:2017,kaiser:depthwise:2017} and transformer networks~\citep{vaswani:transformer:2017}. 
Recent work relies on attention mechanisms where the encoder produces a sequence of vectors and, for each target token, the decoder attends to the most relevant part of the source through a context-dependent weighted-sum of the encoder vectors~\citep{bahdanau:attention:2015,luong:nmt:2015}. Attention has been refined with multi-hop attention~\citep{gehring:convs2s:2017}, self-attention~\citep{vaswani:transformer:2017,paulus:summary:2018} and multi-head attention~\citep{vaswani:transformer:2017}.
We use a transformer architecture \citep{vaswani:transformer:2017}.

\subsection{Semi-supervised NMT}

Monolingual target data has been used to improve the fluency of machine translations since the early IBM models~\citep{brown:smt:1990}. In phrase-based systems, language models (LM) in the target language increase the score of fluent outputs during decoding~\citep{koehn:smt:2003,brants:largelm:2007}. A similar strategy can be applied to NMT~\citep{he:nmt_smt:2016}. Besides improving accuracy during decoding, neural LM and NMT can benefit from deeper integration, e.g. by combining the hidden states of both models~\citep{gulcehre:lm:2017}. 
Neural architecture also allows multi-task learning and parameter sharing between MT and target-side LM~\citep{domhan:mtlm:2017}.

Back-translation (BT) is an alternative to leverage monolingual data. BT is simple and easy to apply as it does not require modification to the MT training algorithms. It requires training a target-to-source system in order to generate additional synthetic parallel data from the monolingual target data. 
This data complements human bitext to train the desired source-to-target system. BT has been applied earlier to phrase-base systems~\citep{bojar:bt_pbmt:2011}. For these systems, BT has also been successful in leveraging monolingual data for domain adaptation~\cite{bertoldi:adaptation:2009,lambert:adaptation:2011}. Recently, BT has
been shown beneficial for NMT~\citep{sennrich:backtranslation:2016,poncelas:investigatingBT:2018}. It has been found to be particularly useful when parallel data is scarce~\citep{karakanta:lowresource:2017}.

\citet{currey:copybt:2017} show that low resource language pairs can also be improved with synthetic data where the source is simply a copy of the monolingual target data.
Concurrently to our work, \citet{imamura:enhancement:2018} show that sampling synthetic sources is more effective than beam search. 
Specifically, they sample multiple sources for each target whereas we draw only a single sample, opting to train on a larger number of target sentences instead.
\citet{hoang:iterative:2018} and \citet{cotterell:explaining:2018} suggest an iterative procedure which continuously improves the quality of the back-translation and final systems.
\citet{niu:bi:2018} experiment with a multilingual model that does both the forward and backward translation which is continuously trained with new synthetic data.

There has also been work using source-side monolingual data~\cite{zhang:sourcemono:2016}. 
Furthermore, \citet{cheng:semi:2016,he:dual:2016,xia:dual:2017} show how monolingual text from both languages can be leveraged by extending back-translation to dual learning: when training both source-to-target and target-to-source models jointly, one can use back-translation in both directions and perform multiple rounds of BT. A similar idea is applied in unsupervised NMT~\citep{lample:unsupervised:2018,lample:unsupervised2:2018}.
Besides monolingual data, various approaches have been introduced to benefit from parallel data in other language pairs~\citep{johnson:multi:2017,firat:multi:2016,firat:zero:2016,ha:multi:2016,gu:universal:2018}.

Data augmentation is an established technique in computer vision where a labeled dataset is supplemented with cropped or rotated input images. 
Recently, generative adversarial networks (GANs) have been successfully used to the same end ~\citep{antoniou:augmentation:2017,perez:augmentation:2017} as well as models that learn distributions over image transformations~\citep{hauberg:dreaming:2016}.

\section{Generating synthetic sources}
\label{sec:genmethod}

Back-translation typically uses beam search \citep{sennrich:backtranslation:2016} or just greedy search \citep{lample:unsupervised:2018,lample:unsupervised2:2018} to generate synthetic source sentences.
Both are approximate algorithms to identify the maximum a-posteriori (MAP) output, i.e. the sentence with the largest estimated probability given an input.
Beam is generally successful in finding high probability outputs~\citep{ott:uncertainty:2018}.

However, MAP prediction can lead to less rich translations~\citep{ott:uncertainty:2018} since it always favors the most likely alternative in case of ambiguity. 
This is particularly problematic in tasks where there is a high level of uncertainty such as dialog~\cite{serban:dialog:2016} and story generation~\cite{fan:stories:2018}. 
We argue that this is also problematic for a data augmentation scheme such as back-translation. 
Beam and greedy focus on the head of the model distribution which results in very regular synthetic source sentences that do not properly cover the true data distribution.

As alternative, we consider sampling from the model distribution as well as adding noise to beam search outputs.
First, we explore unrestricted sampling which generates outputs that are very diverse but sometimes highly unlikely. 
Second, we investigate sampling restricted to the most likely words~\citep{graves:generating:2013,ott:uncertainty:2018,fan:stories:2018}. 
At each time step, we select the $k$ most likely tokens from the output distribution, re-normalize and then sample from this restricted set. 
This is a middle ground between MAP and unrestricted sampling.

As a third alternative, we apply noising \citet{lample:unsupervised:2018} to beam search outputs. 
Adding noise to input sentences has been very beneficial for the autoencoder setups of \citep{lample:unsupervised:2018,hill:naacl:2016} which is inspired by denoising autoencoders \citep{vincent:2008:icml}.
In particular, we transform source sentences with three types of noise: deleting words with probability 0.1, replacing words by a filler token with probability 0.1, and swapping words which is implemented as a random permutation over the tokens, drawn from the uniform distribution but restricted to swapping words no further than three positions apart.

\section{Experimental setup}
\label{sec:setup}

\subsection{Datasets}

The majority of our experiments are based on data from the WMT'18 English-German news translation task. 
We train on all available bitext excluding the ParaCrawl corpus and remove sentences longer than 250 words as well as sentence-pairs with a source/target length ratio exceeding 1.5.
This results in 5.18M sentence pairs. 
For the back-translation experiments we use the German monolingual newscrawl data distributed with WMT'18 comprising 226M sentences after removing duplicates.
We tokenize all data with the Moses tokenizer~\cite{koehn:moses:2007} and learn a joint source and target Byte-Pair-Encoding (BPE; Sennrich et al., 2016)\nocite{sennrich:bpe:2016} with 35K types.
We develop on newstest2012 and report final results on newstest2013-2017; additionally we consider a held-out set from the training data of 52K sentence-pairs.

We also experiment on the larger WMT'14 English-French task which we filter in the same way as WMT'18 English-German.
This results in 35.7M sentence-pairs for training and we learn a joint BPE vocabulary of 44K types.
As monolingual data we use newscrawl2010-2014, comprising 31M sentences after language identification \citep{lui2012langid}.
We use newstest2012 as development set and report final results on newstest2013-2015.

The majority of results in this paper are in terms of case-sensitive tokenized BLEU~\cite{papineni:bleu:2002} but we also report test accuracy with de-tokenized BLEU using sacreBLEU \citep{post:sacre:2018}.

\subsection{Model and hyperparameters}

We re-implemented the Transformer model in pytorch using the fairseq toolkit.\footnote{Code available at \\ \url{https://github.com/pytorch/fairseq}} 
All experiments are based on the Big Transformer architecture with 6 blocks in the encoder and decoder. 
We use the same hyper-parameters for all experiments, i.e., word representations of size 1024, feed-forward layers with inner dimension 4096. 
Dropout is set to 0.3 for En-De and 0.1 for En-Fr, we use 16 attention heads, and we average the checkpoints of the last ten epochs.
Models are optimized with Adam \citep{kingma:adam:2015} using $\beta_1 = 0.9$, $\beta_2 = 0.98$, and $\epsilon = 1e-8$ and we use the same learning rate schedule as \citet{vaswani:transformer:2017}.
All models use label smoothing with a uniform prior distribution over the vocabulary $\epsilon = 0.1$ \citep{szegedy:inception:2015,pereyra:regularize:2017}. 
We run experiments on DGX-1 machines with 8 Nvidia V100 GPUs and machines are interconnected by Infiniband.
Experiments are run on 16 machines and we perform 30K synchronous updates.
We also use the NCCL2 library and the torch distributed package for inter-GPU communication.
We train models with 16-bit floating point operations, following~\citet{ott:scaling:2018}.
For final evaluation, we generate translations with a beam of size 5 and with no length penalty.

\section{Results}
\label{sec:result}

Our evaluation first compares the accuracy of back-translation generation methods (\textsection\ref{sec:genmethod_exp}) and analyzes the results (\textsection\ref{sec:analysis}). 
Next, we simulate a low-resource setup to experiment further with different generation methods (\textsection\ref{sec:lowres}).
We also compare synthetic bitext to genuine parallel data and examine domain effects arising in back-translation (\textsection\ref{sec:domain}). 
We also measure the effect of upsampling bitext during training (\textsection\ref{sec:upsample}). 
Finally, we scale to a very large setup of up to 226M monolingual sentences and compare to previous research (\textsection\ref{sec:largescale}).

\subsection{Synthetic data generation methods}
\label{sec:genmethod_exp}

We first investigate different methods to generate synthetic source translations given a back-translation model, i.e., a model trained in the reverse language direction (Section~\ref{sec:genmethod}).
We consider two types of MAP prediction: greedy search (greedy) and beam search with beam size $5$ (beam). 
Non-MAP methods include unrestricted sampling from the model distribution (sampling), restricting sampling to the $k$ highest scoring outputs at every time step with $k=10$ (top10) as well as adding noise to the beam outputs (beam+noise).
Restricted sampling is a middle-ground between beam search and unrestricted sampling, it is less likely to pick very low scoring outputs but still preserves some randomness. Preliminary experiments with top5, top20, top50 gave similar results to top10.

We also vary the amount of synthetic data and perform 30K updates during training for the bitext only, 50K updates when adding 3M synthetic sentences, 75K updates for 6M and 12M sentences and 100K updates for 24M sentences.
For each setting, this corresponds to enough updates to reach convergence in terms of held-out loss. 
In our 128 GPU setup, training of the final models takes 3h 20min for the bitext only model, 7h 30min for 6M and 12M synthetic sentences, and 10h 15min for 24M sentences. 
During training we also sample the bitext more frequently than the synthetic data and we analyze the effect of this in more detail in \textsection\ref{sec:upsample}.

\begin{figure}[t]
\begin{center}
\resizebox {1\columnwidth}{!}{\begin{tikzpicture}
\begin{axis}[
  xlabel=Total training data,
  ylabel=BLEU (newstest2012),
  xtick={0,1,2,3,4},
  xticklabels={5M,8M,11M,17M,29M},
  xmin=0,
  xmax=4,
  style={thick},
  legend style={at={(1,0.14)},anchor=east,font=\normalsize},
 legend columns=2, 
  grid=both]
\addplot table [y=greedy, x=data]{gen_newstest2012_upsample.dat};
\addlegendentry{greedy}
\addplot table [y=beam, x=data]{gen_newstest2012_upsample.dat};
\addlegendentry{beam}
\addplot+[mark=x] table [y=top10, x=data]{gen_newstest2012_upsample.dat};
\addlegendentry{top10}
\addplot table [y=sampling, x=data]{gen_newstest2012_upsample.dat};
\addlegendentry{sampling}
\addplot[color=orange,mark=triangle] table [y=beam+noise, x=data]{gen_newstest2012_upsample.dat};
\addlegendentry{beam+noise}
\end{axis}
\end{tikzpicture}}
\caption{Accuracy of models trained on different amounts of back-translated data obtained with greedy search, beam search ($k=5$), randomly sampling from the model distribution, restricting sampling over the ten most likely words (top10), and by adding noise to the beam outputs (beam+noise). Results based on newstest2012 of WMT English-German translation.
\label{fig:genmethod}}
\end{center}
\end{figure}
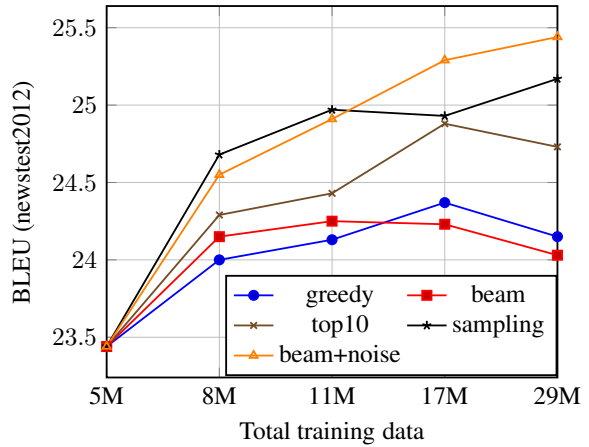

\begin{table*}[t]
\centering
\begin{tabular}{lrrrrrrr}
\toprule
& \bf \bf news2013 & \bf news2014 & \bf news2015 & \bf news2016 & \bf news2017 & \bf Average \\ \midrule
bitext & 27.84 & 30.88 & 31.82 & 34.98 & 29.46 & 31.00 \\ \midrule
+ beam & 27.82 & 32.33 & 32.20 & 35.43 & 31.11 & 31.78 \\ 
+ greedy & 27.67 & 32.55 & 32.57 & 35.74 & 31.25 & 31.96 \\ 
+ top10 & 28.25 & 33.94 & 34.00 & 36.45 & 32.08 & 32.94 \\ 
+ sampling & 28.81 & 34.46 & 34.87 & 37.08 & 32.35 & 33.51 \\ 
+ beam+noise & 29.28 & 33.53 & 33.79 & 37.89 & 32.66 & 33.43 \\ 
\bottomrule
\end{tabular}
\caption{Tokenized BLEU on various test sets of WMT English-German when adding 24M synthetic sentence pairs obtained by various generation methods to a 5.2M sentence-pair bitext (cf. Figure~\ref{fig:genmethod}).
}
\label{tab:gentest_detok}
\end{table*}

Figure~\ref{fig:genmethod} shows that sampling and beam+noise outperform the MAP methods (pure beam search and greedy) by 0.8-1.1 BLEU.
Sampling and beam+noise improve over bitext-only (5M) by between 1.7-2 BLEU in the largest data setting.
Restricted sampling (top10) performs better than beam and greedy but is not as effective as unrestricted sampling (sampling) or beam+noise.

Table~\ref{tab:gentest_detok} shows results on a wider range of test sets (newstest2013-2017). 
Sampling and beam+noise perform roughly equal and we adopt sampling for the remaining experiments.

\subsection{Analysis of generation methods}
\label{sec:analysis}

\begin{figure}[t]
\begin{center}
\resizebox {1\columnwidth}{!}{\begin{tikzpicture}
\begin{axis}[
  xmin=1,
  xmax=100,
  xlabel=epoch,
  ylabel=Training perplexity,
  ymax=6,
  mark repeat=10,
  legend columns=2, 
  legend style={at={(0.34,0.67)},anchor=west,font=\small},
  xtick={1,20,40,60,80,100},
]
\addplot table [y=greedy, x=epoch]{ppl_bt.dat};
\addlegendentry{greedy}
\addplot table [y=beam, x=epoch]{ppl_bt.dat};
\addlegendentry{beam}
\addplot table [y=top10, x=epoch]{ppl_bt.dat};
\addlegendentry{top10}
\addplot table [y=sampling, x=epoch]{ppl_bt.dat};
\addlegendentry{sampling}
\addplot[color=orange,mark=triangle] table [y=beam_noise, x=epoch]{ppl_bt.dat};
\addlegendentry{beam+noise}
\addplot+[no markers] table [y=avg, x=epoch]{ppl_bitext.dat};
\addlegendentry{bitext}
\end{axis}
\end{tikzpicture}}
\caption{Training perplexity (PPL) per epoch for different synthetic data. We separately report PPL on the synthetic data and the bitext. Bitext PPL is averaged over all generation methods.
\label{fig:loss}}
\end{center}
\end{figure}
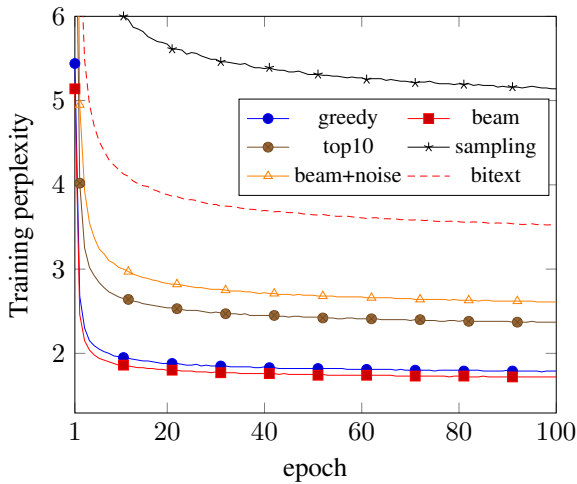

\begin{table}[t]
\centering
\begin{tabular}{lrr}
\toprule
& \bf Perplexity \\ \midrule
human data & 75.34 \\
beam & 72.42 \\ 
sampling & 500.17 \\
top10 & 87.15\\
beam+noise & 2823.73 \\
\bottomrule
\end{tabular}
\caption{Perplexity of source data as assigned by a language model (5-gram Kneser--Ney). Data generated by beam search is most predictable.\\
}
\label{tab:kenlm}
\end{table}

\begin{table*}[t]
\centering
\begin{tabular}{l | p{0.85\textwidth}}
\toprule
source & Diese gegensätzlichen Auffassungen von Fairness liegen nicht nur der politischen Debatte zugrunde. \\
\grow{reference} & These competing principles of fairness underlie not only the political debate. \\
beam & These conflicting interpretations of fairness are not solely based on the political debate. \\
\grow{sample} & \emph{Mr President,} these contradictory interpretations of fairness are not based solely on the political debate. \\
top10 & Those conflicting interpretations of fairness are not solely at the heart of the political debate. \\
\grow{beam+noise} & conflicting BLANK interpretations BLANK are of not BLANK based on the political debate. \\
\bottomrule
\end{tabular}
\caption{Example where sampling produces inadequate outputs. "Mr President," is not in the source. BLANK means that a word has been replaced by a filler token.
}
\label{tab:sample_vs_topk}
\end{table*}

The previous experiment showed that synthetic source sentences generated via sampling and beam with noise perform significantly better than those obtained by pure MAP methods.
Why is this?

Beam search focuses on very likely outputs which reduces the diversity and richness of the generated source translations. 
Adding noise to beam outputs and sampling do not have this problem:
Noisy source sentences make it harder to predict the target translations which may help learning, similar to denoising autoencoders \citep{vincent:2008:icml}.
Sampling is known to better approximate the data distribution which is richer than the argmax model outputs \citep{ott:uncertainty:2018}. 
Therefore, sampling is also more likely to provide a richer training signal than argmax sequences. 

To get a better sense of the training signal provided by each method, we compare the loss on the training data for each method.
We report the cross entropy loss averaged over all tokens and separate the loss over the synthetic data and the real bitext data.
Specifically, we choose the setup with 24M synthetic sentences. At the end of each epoch we measure the loss over 500K sentence pairs sub-sampled from the synthetic data as well as an equally sized subset of the bitext. 
For each generation method we choose the same sentences except for the bitext which is disjoint from the synthetic data.
This means that losses over the synthetic data are measured over the \emph{same} target tokens because the generation methods only differ in the source sentences.
We found it helpful to upsample the frequency with which we observe the bitext compared to the synthetic data (\textsection\ref{sec:upsample}) but we do not upsample for this experiment to keep conditions as similar as possible.
We assume that when the training loss is low, then the model can easily fit the training data without extracting much learning signal compared to data which is harder to fit.

Figure~\ref{fig:loss} shows that synthetic data based on greedy or beam is much easier to fit compared to data from sampling, top10, beam+noise and the bitext.
In fact, the perplexity on beam data falls below 2 after only 5 epochs.
Except for sampling, we find that the perplexity on the training data is somewhat correlated to the end-model accuracy (cf. Figure~\ref{fig:genmethod}) and that all methods except sampling have a lower loss than real bitext.

These results suggest that synthetic data obtained with argmax inference does not provide as rich a training signal as sampling or adding noise. 
We conjecture that the regularity of synthetic data obtained with argmax inference is not optimal.
Sampling and noised argmax both expose the model to a wider range of source sentences which makes the model more robust to reordering and substitutions that happen naturally, even if the model of reordering and substitution through noising is not very realistic.

Next we analyze the richness of synthetic outputs and train a language model on real human text and score synthetic source sentences generated by beam search, sampling, top10 and beam+noise.
We hypothesize that data that is very regular should be more predictable by the language model and therefore receive low perplexity. 
We eliminate a possible domain mismatch effect between the language model training data and the synthetic data by splitting the parallel corpus into three non-overlapping parts:
\begin{enumerate}
\item On 640K sentences pairs, we train a back-translation model,
\item On 4.1M sentence pairs, we take the source side and train a 5-gram Kneser-Ney language model~\citep{heafield:acl:2013}, 
\item On the remaining 450K sentences, we apply the back-translation system using beam, sampling and top10 generation. 
\end{enumerate}

For the last set, we have genuine source sentences as well as synthetic sources from different generation techniques. 
We report the perplexity of our language model on all versions of the source data in Table~\ref{tab:kenlm}. 
The results show that beam outputs receive higher probability by the language model compared to sampling, beam+noise and real source sentences. 
This indicates that beam search outputs are not as rich as sampling outputs or beam+noise. 
This lack of variability probably explains in part why back-translations from pure beam search provide a weaker training signal than alternatives.

Closer inspection of the synthetic sources (Table~\ref{tab:sample_vs_topk}) reveals that sampled and noised beam outputs are sometimes not very adequate, much more so than MAP outputs, e.g., sampling often introduces target words which have no counterpart in the source.
This happens because sampling sometimes picks highly unlikely outputs which are harder to fit (cf. Figure~\ref{fig:loss}).

\subsection{Low resource vs. high resource setup}
\label{sec:lowres}

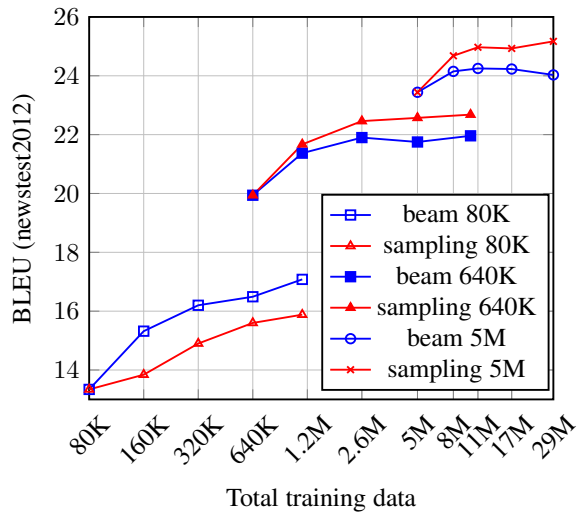
\begin{figure}[t]
\begin{center}
\resizebox {1\columnwidth}{!}{\begin{tikzpicture}
\begin{axis}[
   xlabel=Total training data,
  xmode=log,
  ylabel=BLEU (newstest2012),
  ymin=13,
  ymax=26,
  xtick={80,160,320,640,1280,2560,5190,8190,11190,17190,29190},
  xticklabels={80K,160K,320K,640K,1.2M,2.6M,5M,8M,11M,17M,29M},
  xticklabel style=
		{rotate=45,anchor=near xticklabel},
  xmin=80,
  xmax=29190,
  style={thick},
  legend style={at={(0.5,0.28)},anchor=west},
  grid=both]
\addplot[color=blue,mark=square] table [y=beam_80, x=data]
{beam_vs_sample_3.dat};
\addlegendentry{beam 80K}
\addplot[color=red,mark=triangle] table [y=sample_80, x=data]{beam_vs_sample_3.dat};
\addlegendentry{sampling 80K}
\addplot[color=blue,mark=square*] table [y=beam_640, x=data]{beam_vs_sample_3.dat};
\addlegendentry{beam 640K}
\addplot[color=red,mark=triangle*] table [y=sample_640, x=data]{beam_vs_sample_3.dat};
\addlegendentry{sampling 640K}
\addplot[color=blue,mark=o] table [y=beam_5M, x=data]{beam_vs_sample_3.dat};
\addlegendentry{beam 5M}
\addplot[color=red,mark=x] table [y=sample_5M, x=data]{beam_vs_sample_3.dat};
\addlegendentry{sampling 5M}
\end{axis}
\end{tikzpicture}}
\caption{BLEU when adding synthetic data from beam and sampling to bitext systems with 80K, 640K and 5M sentence pairs.
\label{fig:beam_vs_sample}}
\end{center}
\end{figure}

The experiments so far are based on a setup with a large bilingual corpus. 
However, in resource poor settings the back-translation model is of much lower quality. 
Are non-MAP methods still more effective in such a setup?
To answer this question, we simulate such setups by sub-sampling the training data to either 80K sentence-pairs or 640K sentence-pairs and then add synthetic data from sampling and beam search. 
We compare these smaller setups to our original 5.2M sentence bitext configuration.
The accuracy of the German-English back-translation systems steadily increases with more training data: On newstest2012 we measure 13.5 BLEU for 80K bitext, 24.3 BLEU for 640K and 28.3 BLEU for 5M.

Figure~\ref{fig:beam_vs_sample} shows that sampling is more effective than beam for larger setups (640K and 5.2M bitexts) while the opposite is true for resource poor settings (80K bitext).
This is likely because the back-translations in the 80K setup are of very poor quality and the noise of sampling and beam+noise is too detrimental for this brittle low-resource setting.
When the setup is very small the very regular MAP outputs still provide useful training signal while the noise from sampling becomes harmful.

\subsection{Domain of synthetic data}
\label{sec:domain}

Next, we turn to two different questions: 
How does real human bitext compare to synthetic data in terms of final model accuracy? 
And how does the domain of the monolingual data affect results?

To answer these questions, we subsample 640K sentence-pairs of the bitext and train a back-translation system on this set. To train a forward model, we consider three alternative types of data to add to this 640K training set. We either add:
\begin{itemize}
\item the remaining parallel data (bitext),
\item the back-translated target side of the remaining parallel data (BT-bitext),
\item back-translated newscrawl data (BT-news).
\end{itemize}
The back-translated data is generated via sampling. 
This setup allows us to compare synthetic data to genuine data since BT-bitext and bitext share the same target side.
It also allows us to estimate the value of BT data for domain adaptation since the newscrawl corpus (BT-news) is pure news whereas the bitext is a mixture of europarl and commoncrawl with only a small news-commentary portion.
To assess domain adaptation effects, we measure accuracy on two held-out sets:
\begin{itemize}
\item newstest2012, i.e. pure newswire data.
\item a held-out set of the WMT training data (valid-mixed), which is a mixture of europarl, commoncrawl and the small news-commentary portion.
\end{itemize}

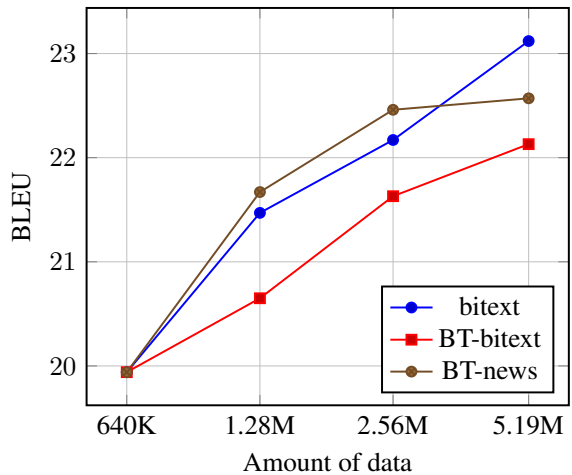
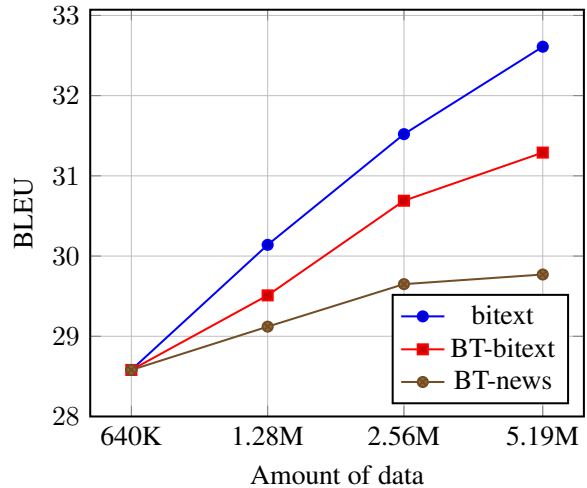
\begin{figure*}[t]
\centering
\begin{subfigure}[b]{0.48\textwidth}
\centering
\resizebox {1\columnwidth}{!}{
\begin{tikzpicture}
\begin{axis}[
  xmode=log,
  xlabel=Amount of data,
  ylabel=BLEU,
  xtick={640,1280,2560,5186},
  xticklabels={640K,1.28M,2.56M,5.19M},
  style={thick},
  legend pos=south east,
  grid=both]
\addplot table [y=real, x=data]{domain_newstest2012.dat};
\addlegendentry{bitext}
\addplot table [y=btbitext, x=data]{domain_newstest2012.dat};
\addlegendentry{BT-bitext}
\addplot table [y=btnews, x=data]{domain_newstest2012.dat};
\addlegendentry{BT-news}
\end{axis}
\end{tikzpicture}
}
\caption{newstest2012}
\label{fig:domain_news}
\end{subfigure}
\hfill
\begin{subfigure}[b]{0.48\textwidth}
\resizebox {1\columnwidth}{!}{\begin{tikzpicture}
\begin{axis}[
  xmode=log,
  ymin=28,
  xlabel=Amount of data,
  ylabel=BLEU,
  xtick={640,1280,2560,5186},
  xticklabels={640K,1.28M,2.56M,5.19M},
  style={thick},
  legend pos=south east,
  grid=both]
\addplot table [y=real, x=data]{domain_valid.dat};
\addlegendentry{bitext}
\addplot table [y=btbitext, x=data]{domain_valid.dat};
\addlegendentry{BT-bitext}
\addplot table [y=btnews, x=data]{domain_valid.dat};
\addlegendentry{BT-news}
\end{axis}
\end{tikzpicture}}
\caption{valid-mixed}
\label{fig:domain_valid}
\end{subfigure}
\caption{Accuracy on (\subref{fig:domain_news}) newstest2012 and (\subref{fig:domain_valid}) a mixed domain valid set when growing a 640K bitext corpus with (i) real parallel data (bitext), (ii) a back-translated version of the target side of the bitext (BT-bitext), (iii) or back-translated newscrawl data (BT-news).}
\label{fig:domain}
\end{figure*}

Figure~\ref{fig:domain} shows the results on both validation sets. 
Most strikingly, BT-news performs almost as well as bitext on newstest2012 (Figure~\ref{fig:domain_news}) and improves the baseline (640K) by 2.6 BLEU.
BT-bitext improves by 2.2 BLEU, achieving 83\% of the improvement with real bitext.
This shows that synthetic data can be nearly as effective as real human translated data when the domains match.

Figure~\ref{fig:domain_valid} shows the accuracy on valid-mixed, the mixed domain valid set.
The accuracy of BT-news is not as good as before since the domain of the BT data and the test set do not match.
However, BT-news still improves the baseline by up to 1.2 BLEU.
On the other hand, BT-bitext matches the domain of valid-mixed and improves by 2.7 BLEU.
This trails the real bitext by only 1.3 BLEU and corresponds to 67\% of the gain achieved with real human bitext.

In summary, synthetic data performs remarkably well, coming close to the improvements achieved with real bitext for newswire test data, or trailing real bitext by only 1.3 BLEU for valid-mixed. 
In absence of a large parallel corpus for news, back-translation therefore offers a simple, yet very effective domain adaptation technique.

\subsection{Upsampling the bitext}
\label{sec:upsample}

We found it beneficial to adjust the ratio of bitext to synthetic data observed during training.
In particular, we tuned the rate at which we sample data from the bitext compared to synthetic data. 
For example, in a setup of 5M bitext sentences and 10M synthetic sentences, an upsampling rate of 2 means that we double the frequency at which we visit bitext, i.e. training batches contain on average an equal amount of bitext and synthetic data as opposed to 1/3 bitext and 2/3 synthetic data.

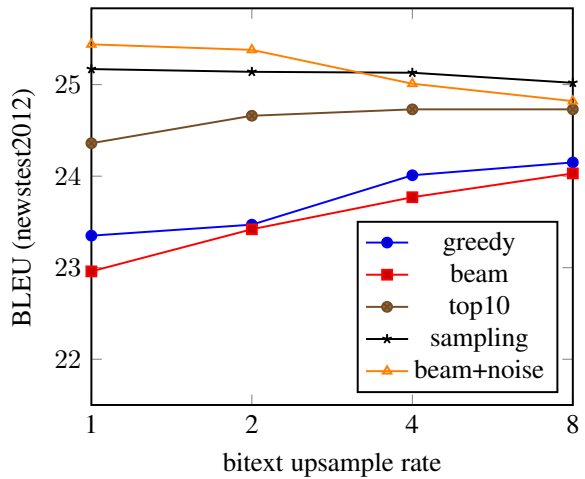
\begin{figure}[t]
\begin{center}
\resizebox {1\columnwidth}{!}{\begin{tikzpicture}
\begin{axis}[
  xmode=log,
  xlabel=bitext upsample rate,
  ylabel=BLEU (newstest2012),
  xmin=1,
  xmax=8,
  ymin=21.5,
  xtick={1,2,4,8},
  xticklabels={1,2,4,8},
  style={thick},
  legend pos=south east,
]
\addplot+ table [y=greedy, x=rate]{upsample.dat};
\addlegendentry{greedy}
\addplot+ table [y=beam, x=rate]{upsample.dat};
\addlegendentry{beam}
\addplot+ table [y=top10, x=rate]{upsample.dat};
\addlegendentry{top10}
\addplot+ table [y=sampling, x=rate]{upsample.dat};
\addlegendentry{sampling}
\addplot+[color=orange, mark=triangle] table [y=beam_noise, x=rate]{upsample.dat};
\addlegendentry{beam+noise}
\end{axis}
\end{tikzpicture}}
\caption{Accuracy when changing the rate at which the bitext is upsampled during training. Rates larger than one mean that the bitext is observed more often than actually present in the combined bitext and synthetic training corpus.
\label{fig:upsample}}
\end{center}
\end{figure}

Figure~\ref{fig:upsample} shows the accuracy of various upsampling rates for different generation methods in a setup with 5M bitext sentences and 24M synthetic sentences.
Beam and greedy benefit a lot from higher rates which results in training more on the bitext data. 
This is likely because synthetic beam and greedy data does not provide as much training signal as the bitext which has more variation and is harder to fit.
On the other hand, sampling and beam+noise require no upsampling of the bitext, which is likely because the synthetic data is already hard enough to fit and thus provides a strong training signal (\textsection\ref{sec:analysis}).

\subsection{Large scale results}
\label{sec:largescale}

To confirm our findings we experiment on WMT'14 English-French translation where we show results on newstest2013-2015. 
We augment the large bitext of 35.7M sentence pairs by 31M newscrawl sentences generated by sampling. 
To train this system we perform 300K training updates in 27h 40min on 128 GPUs; we do not upsample the bitext for this experiment.
Table~\ref{tab:en2fr} shows tokenized BLEU and Table~\ref{tab:en2fr_detok} shows detokenized BLEU.\footnote{sacreBLEU signatures: BLEU+case.mixed+lang.en-fr+numrefs.1+smooth.exp+test.SET+tok.13a+version.1.2.7 with SET $\in$ \{wmt13, wmt14/full, wmt15\}}
To our knowledge, our baseline is the best reported result in the literature for newstest2014, and back-translation further improves upon this by 2.6 BLEU (tokenized).

\begin{table}[t]
\centering
\begin{tabular}{lrrr}
\toprule
& \bf news13 & \bf news14 & \bf news15 \\ \midrule
bitext    & 36.97 & 42.90 & 39.92 \\
+sampling & {\bf 37.85} & {\bf 45.60} & {\bf 43.95} \\
\bottomrule
\end{tabular}
\caption{Tokenized BLEU on various test sets for WMT English-French translation.
}
\label{tab:en2fr}
\end{table}

\begin{table}[t]
\centering
\begin{tabular}{lrrr}
\toprule
& \bf news13 & \bf news14 & \bf news15 \\ \midrule
bitext    & 35.30 & 41.03 & 38.31 \\
+sampling & {\bf 36.13} & {\bf 43.84} & {\bf 40.91}\\
\bottomrule
\end{tabular}
\caption{De-tokenized BLEU (sacreBLEU) on various test sets for WMT English-French.}
\label{tab:en2fr_detok}
\end{table}

\begin{table}[t]
\centering
\begin{tabular}{l|r r}
\toprule
                                & En--De   & En--Fr \\
\hline
a. \citet{gehring:convs2s:2017}    &  25.2    & 40.5 \\
b. \citet{vaswani:transformer:2017}&  28.4    & 41.0 \\
c. \citet{ahmed:weightedtn:2017}   &  28.9    & 41.4 \\
d. \citet{shaw:relpos:2018}        &  29.2    & 41.5 \\
\midrule
DeepL                           &  33.3       & {\bf 45.9} \\
Our result                      & {\bf 35.0}  & 45.6 \\
~~~{\it detok. sacreBLEU\tablefootnote{sacreBLEU signatures: BLEU+case.mixed+lang.en-LANG+numrefs.1+smooth.exp+test.wmt14/full+ tok.13a+version.1.2.7 with LANG $\in$ \{de,fr\}}}      &  {\it 33.8}  & {\it 43.8} \\
\bottomrule
\end{tabular}
\caption{BLEU on newstest2014 for WMT English-German (En--De) and English-French (En--Fr). The first four results use only WMT bitext (WMT'14, except for b, c, d in En--De which train on WMT'16). 
DeepL uses proprietary high-quality bitext and our result relies on back-translation with 226M newscrawl sentences for En--De and 31M for En--Fr. We also show detokenized BLEU (SacreBLEU).
}
\label{tab:testwmt}
\end{table}

Finally, for WMT English-German we train on all 226M available monolingual training sentences and perform 250K updates in 22.5 hours on 128 GPUs. 
We upsample the bitext with a rate of 16 so that we observe every bitext sentence 16 times more often than each monolingual sentence.
This results in a new state of the art of 35 BLEU on newstest2014 by using only WMT benchmark data.
For comparison, DeepL, a commercial translation engine relying on high quality bilingual training data, achieves 33.3 tokenized BLEU .\tablefootnote{\url{https://www.deepl.com/press.html}}
Table~\ref{tab:testwmt} summarizes our results and compares to other work in the literature.
This shows that back-translation with sampling can result in high-quality translation models based on benchmark data only.

\section{Submission to WMT'18}

This section describes our entry to the WMT'18 English-German news translation task which was ranked \#1 in the human evaluation \citep{bojar-EtAl:2018:WMT1}.
Our entry is based on the WMT English-German models described in the previous section (\textsection\ref{sec:largescale}).
In particular, we ensembled six back-translation models trained on all available bitext plus 226M newscrawl sentences or 5.8B German tokens.
Four models used bitext upsample ratio 16, one model upsample ratio 32, and another one upsample ratio 8.
Upsample ratios differed because we reused models previously trained to tune the upsample ratio. 
We did not use checkpoint averaging. 
More details of our setup and data are described in \textsection\ref{sec:setup}.

\citet{ott:uncertainty:2018} showed that beam search sometimes outputs source copies rather than target language translations. 
We replaced source copies by the output of a model trained only on the news-commentary portion of the WMT'18 task (nc model). 
This model produced far fewer copies since this dataset is less noisy.
Outputs are deemed to be a source copy if the Jaccard similarity between the source and the target unigrams exceeds 0.5.
About 0.5\% of outputs are identified as source copies.
We used newstest17 as a development set to fine tune ensemble size and model parameters. 
Table~\ref{tab:en2de_wmt18} summarizes the effect of back-translation data, ensembling and source copy filtering.\tablefootnote{BLEU+case.lc+lang.en-de+numrefs.1+smooth.exp+test.SET+tok.13a+version.1.2.11 with SET $\in$ \{wmt17, wmt18\}}

\begin{table}[t]
\centering
\begin{tabular}{lrrr}
\toprule
& \bf news17 & \bf news18 \\ \midrule
baseline    & 29.36 & 42.38 \\
+BT & 32.66  & 44.94 \\
+ensemble & 33.31 & 46.39 \\
+filter copies & {\bf 33.35} & {\bf 46.53} \\
\midrule
\% of source copies & 0.56\% & 0.53\% \\
\bottomrule
\end{tabular}
\caption{De-tokenized case-insensitive sacreBLEU on WMT English-German newstest17 and newstest18.}
\label{tab:en2de_wmt18}
\end{table}

\section{Conclusions and future work}
\label{sec:ccl}

Back-translation is a very effective data augmentation technique for neural machine translation. 
Generating synthetic sources by sampling or by adding noise to beam outputs leads to higher accuracy than argmax inference which is typically used. 
In particular, sampling and noised beam outperforms pure beam by 1.7 BLEU on average on newstest2013-2017 for WMT English-German translation.
Both methods provide a richer training signal for all but resource poor setups.
We also find that synthetic data can achieve up to 83\% of the performance attainable with real bitext. 
Finally, we achieve a new state of the art result of 35 BLEU on the WMT'14 English-German test set by using publicly available benchmark data only.

In future work, we would like to investigate an end-to-end approach where the back-translation model is optimized to output synthetic sources that are most helpful to the final forward model.

\bibliography{main}

\begin{thebibliography}{62}
\expandafter\ifx\csname natexlab\endcsname\relax\def\natexlab#1{#1}\fi

\bibitem[{Ahmed et~al.(2017)Ahmed, Keskar, and Socher}]{ahmed:weightedtn:2017}
Karim Ahmed, Nitish~Shirish Keskar, and Richard Socher. 2017.
\newblock Weighted transformer network for machine translation.
\newblock \emph{arxiv}, 1711.02132.

\bibitem[{Antoniou et~al.(2017)Antoniou, Storkey, and
  Edwards}]{antoniou:augmentation:2017}
Antreas Antoniou, Amos~J. Storkey, and Harrison Edwards. 2017.
\newblock Data augmentation generative adversarial networks.
\newblock \emph{arXiv}, abs/1711.04340.

\bibitem[{Bahdanau et~al.(2015)Bahdanau, Cho, and
  Bengio}]{bahdanau:attention:2015}
Dzmitry Bahdanau, Kyunghyun Cho, and Yoshua Bengio. 2015.
\newblock Neural machine translation by jointly learning to align and
  translate.
\newblock In \emph{International Conference on Learning Representations
  ({ICLR})}.

\bibitem[{Bertoldi and Federico(2009)}]{bertoldi:adaptation:2009}
Nicola Bertoldi and Marcello Federico. 2009.
\newblock Domain adaptation for statistical machine translation with
  monolingual resources.
\newblock In \emph{Workshop on Statistical Machine Translation ({WMT})}.

\bibitem[{Bojar and Tamchyna(2011)}]{bojar:bt_pbmt:2011}
Ondrej Bojar and Ales Tamchyna. 2011.
\newblock Improving translation model by monolingual data.
\newblock In \emph{Workshop on Statistical Machine Translation ({WMT})}.

\bibitem[{Bojar et~al.(2018)Bojar, Federmann, Fishel, Graham, Haddow, Huck,
  Koehn, and Monz}]{bojar-EtAl:2018:WMT1}
Ond\v{r}ej Bojar, Christian Federmann, Mark Fishel, Yvette Graham, Barry
  Haddow, Matthias Huck, Philipp Koehn, and Christof Monz. 2018.
\newblock Findings of the 2018 conference on machine translation ({WMT18}).
\newblock In \emph{Proceedings of the Third Conference on Machine Translation,
  Volume 2: Shared Task Papers}, Brussels, Belgium. Association for
  Computational Linguistics.

\bibitem[{Brants et~al.(2007)Brants, Popat, Xu, Och, and
  Dean}]{brants:largelm:2007}
Thorsten Brants, Ashok~C. Popat, Peng Xu, Franz~Josef Och, and Jeffrey Dean.
  2007.
\newblock Large language models in machine translation.
\newblock In \emph{Conference on Natural Language Learning ({CoNLL})}.

\bibitem[{Brown et~al.(1990)Brown, Cocke, Pietra, Pietra, Jelinek, Lafferty,
  Mercer, and Roossin}]{brown:smt:1990}
Peter~F. Brown, John Cocke, Stephen~Della Pietra, Vincent J.~Della Pietra,
  Frederick Jelinek, John~D. Lafferty, Robert~L. Mercer, and Paul~S. Roossin.
  1990.
\newblock A statistical approach to machine translation.
\newblock \emph{Computational Linguistics}, 16:79--85.

\bibitem[{Cheng et~al.(2016)Cheng, Xu, He, He, Wu, Sun, and
  Liu}]{cheng:semi:2016}
Yong Cheng, Wei Xu, Zhongjun He, Wei He, Hua Wu, Maosong Sun, and Yang Liu.
  2016.
\newblock Semi-supervised learning for neural machine translation.
\newblock In \emph{Conference of the Association for Computational Linguistics
  ({ACL})}.

\bibitem[{Cho et~al.(2014)Cho, van Merrienboer, Gulcehre, Bahdanau, Bougares,
  Schwenk, and Bengio}]{cho:seq2seq:2014}
Kyunghyun Cho, Bart van Merrienboer, Caglar Gulcehre, Dzmitry Bahdanau, Fethi
  Bougares, Holger Schwenk, and Yoshua Bengio. 2014.
\newblock Learning phrase representations using rnn encoder-decoder for
  statistical machine translation.
\newblock In \emph{Conference on Empirical Methods in Natural Language
  Processing ({EMNLP})}.

\bibitem[{Cotterell and Kreutzer(2018)}]{cotterell:explaining:2018}
Ryan Cotterell and Julia Kreutzer. 2018.
\newblock Explaining and generalizing back-translation through wake-sleep.
\newblock \emph{arXiv preprint arXiv:1806.04402}.

\bibitem[{Currey et~al.(2017)Currey, Barone, and Heafield}]{currey:copybt:2017}
Anna Currey, Antonio Valerio~Miceli Barone, and Kenneth Heafield. 2017.
\newblock {Copied Monolingual Data Improves Low-Resource Neural Machine
  Translation}.
\newblock In \emph{Proc. of WMT}.

\bibitem[{Domhan and Hieber(2017)}]{domhan:mtlm:2017}
Tobias Domhan and Felix Hieber. 2017.
\newblock Using target-side monolingual data for neural machine translation
  through multi-task learning.
\newblock In \emph{Conference on Empirical Methods in Natural Language
  Processing ({EMNLP})}.

\bibitem[{Fan et~al.(2018)Fan, Dauphin, and Lewis}]{fan:stories:2018}
Angela Fan, Yann Dauphin, and Mike Lewis. 2018.
\newblock Hierarchical neural story generation.
\newblock In \emph{Conference of the Association for Computational Linguistics
  ({ACL})}.

\bibitem[{Firat et~al.(2016{\natexlab{a}})Firat, Cho, and
  Bengio}]{firat:multi:2016}
Orhan Firat, Kyunghyun Cho, and Yoshua Bengio. 2016{\natexlab{a}}.
\newblock Multi-way, multilingual neural machine translation with a shared
  attention mechanism.
\newblock In \emph{Conference of the North American Chapter of the Association
  for Computational Linguistics ({NAACL})}.

\bibitem[{Firat et~al.(2016{\natexlab{b}})Firat, Sankaran, Al-Onaizan,
  Yarman-Vural, and Cho}]{firat:zero:2016}
Orhan Firat, Baskaran Sankaran, Yaser Al-Onaizan, Fatos~T. Yarman-Vural, and
  Kyunghyun Cho. 2016{\natexlab{b}}.
\newblock Zero-resource translation with multi-lingual neural machine
  translation.
\newblock In \emph{Conference on Empirical Methods in Natural Language
  Processing ({EMNLP})}.

\bibitem[{Gehring et~al.(2017)Gehring, Auli, Grangier, Yarats, and
  Dauphin}]{gehring:convs2s:2017}
Jonas Gehring, Michael Auli, David Grangier, Denis Yarats, and Yann~N Dauphin.
  2017.
\newblock Convolutional sequence to sequence learning.
\newblock In \emph{International Conference of Machine Learning {(ICML)}}.

\bibitem[{Graves(2013)}]{graves:generating:2013}
Alex Graves. 2013.
\newblock Generating sequences with recurrent neural networks.
\newblock \emph{arXiv}, 1308.0850.

\bibitem[{Gu et~al.(2018)Gu, Hassan, Devlin, and Li}]{gu:universal:2018}
Jiatao Gu, Hany Hassan, Jacob Devlin, and Victor O.~K. Li. 2018.
\newblock Universal neural machine translation for extremely low resource
  languages.
\newblock \emph{arXiv}, 1802.05368.

\bibitem[{Gulcehre et~al.(2015)Gulcehre, Firat, Xu, Cho, Barrault, Lin,
  Bougares, Schwenk, and Bengio}]{gulcehre:lm:2015}
Caglar Gulcehre, Orhan Firat, Kelvin Xu, Kyunghyun Cho, Loic Barrault, Huei-Chi
  Lin, Fethi Bougares, Holger Schwenk, and Yoshua Bengio. 2015.
\newblock On using monolingual corpora in neural machine translation.
\newblock \emph{arXiv}, 1503.03535.

\bibitem[{Gulcehre et~al.(2017)Gulcehre, Firat, Xu, Cho, and
  Bengio}]{gulcehre:lm:2017}
Caglar Gulcehre, Orhan Firat, Kelvin Xu, Kyunghyun Cho, and Yoshua Bengio.
  2017.
\newblock On integrating a language model into neural machine translation.
\newblock \emph{Computer Speech \& Language}, 45:137--148.

\bibitem[{Ha et~al.(2016)Ha, Niehues, and Waibel}]{ha:multi:2016}
Thanh-Le Ha, Jan Niehues, and Alexander~H. Waibel. 2016.
\newblock Toward multilingual neural machine translation with universal encoder
  and decoder.
\newblock \emph{arXiv}, 1611.04798.

\bibitem[{Hassan et~al.(2018)Hassan, Aue, Chen, Chowdhary, Clark, Federmann,
  Huang, Junczys-Dowmunt, Lewis, Li et~al.}]{hassan:parity:2018}
Hany Hassan, Anthony Aue, Chang Chen, Vishal Chowdhary, Jonathan Clark,
  Christian Federmann, Xuedong Huang, Marcin Junczys-Dowmunt, William Lewis,
  Mu~Li, et~al. 2018.
\newblock Achieving human parity on automatic chinese to english news
  translation.
\newblock \emph{arXiv}, 1803.05567.

\bibitem[{Hauberg et~al.(2016)Hauberg, Freifeld, Larsen, Fisher, and
  Hansen}]{hauberg:dreaming:2016}
Soren Hauberg, Oren Freifeld, Anders Boesen~Lindbo Larsen, John~W. Fisher, and
  Lars~Kai Hansen. 2016.
\newblock Dreaming more data: Class-dependent distributions over
  diffeomorphisms for learned data augmentation.
\newblock In \emph{AISTATS}.

\bibitem[{He et~al.(2016{\natexlab{a}})He, Xia, Qin, Wang, Yu, Liu, and
  Ma}]{he:dual:2016}
Di~He, Yingce Xia, Tao Qin, Liwei Wang, Nenghai Yu, Tieyan Liu, and Wei-Ying
  Ma. 2016{\natexlab{a}}.
\newblock Dual learning for machine translation.
\newblock In \emph{Conference on Advances in Neural Information Processing
  Systems ({NIPS})}.

\bibitem[{He et~al.(2016{\natexlab{b}})He, He, Wu, and Wang}]{he:nmt_smt:2016}
Wei He, Zhongjun He, Hua Wu, and Haifeng Wang. 2016{\natexlab{b}}.
\newblock Improved neural machine translation with smt features.
\newblock In \emph{Conference of the Association for the Advancement of
  Artificial Intelligence ({AAAI})}, pages 151--157.

\bibitem[{Heafield et~al.(2013)Heafield, Pouzyrevsky, Clark, and
  Koehn}]{heafield:acl:2013}
Kenneth Heafield, Ivan Pouzyrevsky, Jonathan~H. Clark, and Philipp Koehn. 2013.
\newblock {Scalable Modified {Kneser-Ney} Language Model Estimation}.
\newblock In \emph{Conference of the Association for Computational Linguistics
  ({ACL})}.

\bibitem[{Hill et~al.(2016)Hill, Cho, and Korhonen}]{hill:naacl:2016}
Felix Hill, Kyunghyun Cho, and Anna Korhonen. 2016.
\newblock Learning distributed representations of sentences from unlabelled
  data.
\newblock In \emph{Conference of the North American Chapter of the Association
  for Computational Linguistics ({NAACL})}.

\bibitem[{Hoang et~al.(2018)Hoang, Koehn, Haffari, and
  Cohn}]{hoang:iterative:2018}
Vu~Cong~Duy Hoang, Philipp Koehn, Gholamreza Haffari, and Trevor Cohn. 2018.
\newblock Iterative back-translation for neural machine translation.
\newblock In \emph{Proceedings of the 2nd Workshop on Neural Machine
  Translation and Generation}, pages 18--24.

\bibitem[{Imamura et~al.(2018)Imamura, Fujita, and
  Sumita}]{imamura:enhancement:2018}
Kenji Imamura, Atsushi Fujita, and Eiichiro Sumita. 2018.
\newblock Enhancement of encoder and attention using target monolingual corpora
  in neural machine translation.
\newblock In \emph{Proceedings of the 2nd Workshop on Neural Machine
  Translation and Generation}, pages 55--63.

\bibitem[{Johnson et~al.(2017)Johnson, Schuster, Le, Krikun, Wu, Chen, Thorat,
  Vi{\'e}gas, Wattenberg, Corrado, Hughes, and Dean}]{johnson:multi:2017}
Melvin Johnson, Mike Schuster, Quoc~V. Le, Maxim Krikun, Yonghui Wu, Zhifeng
  Chen, Nikhil Thorat, Fernanda~B. Vi{\'e}gas, Martin Wattenberg, Gregory~S.
  Corrado, Macduff Hughes, and Jeffrey Dean. 2017.
\newblock Google's multilingual neural machine translation system: Enabling
  zero-shot translation.
\newblock \emph{Transactions of the Association for Computational Linguistics
  ({TACL})}, 5:339--351.

\bibitem[{Kaiser et~al.(2017)Kaiser, Gomez, and
  Chollet}]{kaiser:depthwise:2017}
Lukasz Kaiser, Aidan~N. Gomez, and Fran{\c{c}}ois Chollet. 2017.
\newblock Depthwise separable convolutions for neural machine translation.
\newblock \emph{CoRR}, abs/1706.03059.

\bibitem[{Kalchbrenner et~al.(2016)Kalchbrenner, Espeholt, Simonyan, van~den
  Oord, Graves, and Kavukcuoglu}]{kalchbrenner:bytenet:2016}
Nal Kalchbrenner, Lasse Espeholt, Karen Simonyan, A{\"{a}}ron van~den Oord,
  Alex Graves, and Koray Kavukcuoglu. 2016.
\newblock Neural machine translation in linear time.
\newblock \emph{CoRR}, abs/1610.10099.

\bibitem[{Karakanta et~al.(2017)Karakanta, Dehdari, and van
  Genabith}]{karakanta:lowresource:2017}
Alina Karakanta, Jon Dehdari, and Josef van Genabith. 2017.
\newblock Neural machine translation for low-resource languages without
  parallel corpora.
\newblock \emph{Machine Translation}, pages 1--23.

\bibitem[{Kingma and Ba(2015)}]{kingma:adam:2015}
Diederik~P. Kingma and Jimmy Ba. 2015.
\newblock {Adam: A Method for Stochastic Optimization}.
\newblock In \emph{International Conference on Learning Representations
  ({ICLR})}.

\bibitem[{Koehn(2010)}]{koehn:book}
Philipp Koehn. 2010.
\newblock \emph{Statistical machine translation}.
\newblock Cambridge University Press.

\bibitem[{Koehn et~al.(2007)Koehn, Hoang, Birch, Callison-Burch, Federico,
  Bertoldi, Cowan, Shen, Moran, Zens, Dyer, Bojar, Constantin, and
  Herbst}]{koehn:moses:2007}
Philipp Koehn, Hieu Hoang, Alexandra Birch, Chris Callison-Burch, Marcello
  Federico, Nicola Bertoldi, Brooke Cowan, Wade Shen, Christine Moran, Richard
  Zens, Chris Dyer, Ondrej Bojar, Alexandra Constantin, and Evan Herbst. 2007.
\newblock Moses: Open source toolkit for statistical machine translation.
\newblock In \emph{ACL Demo Session}.

\bibitem[{Koehn et~al.(2003)Koehn, Och, and Marcu}]{koehn:smt:2003}
Philipp Koehn, Franz~Josef Och, and Daniel Marcu. 2003.
\newblock Statistical phrase-based translation.
\newblock In \emph{Conference of the North American Chapter of the Association
  for Computational Linguistics ({NAACL})}.

\bibitem[{Lambert et~al.(2011)Lambert, Schwenk, Servan, and
  Abdul-Rauf}]{lambert:adaptation:2011}
Patrik Lambert, Holger Schwenk, Christophe Servan, and Sadaf Abdul-Rauf. 2011.
\newblock Investigations on translation model adaptation using monolingual
  data.
\newblock In \emph{Workshop on Statistical Machine Translation ({WMT})}.

\bibitem[{Lample et~al.(2018{\natexlab{a}})Lample, Conneau, Denoyer, and
  Ranzato}]{lample:unsupervised:2018}
Guillaume Lample, Alexis Conneau, Ludovic Denoyer, and Marc'Aurelio Ranzato.
  2018{\natexlab{a}}.
\newblock Unsupervised machine translation using monolingual corpora only.
\newblock In \emph{International Conference on Learning Representations
  ({ICLR})}.

\bibitem[{Lample et~al.(2018{\natexlab{b}})Lample, Ott, Conneau, Denoyer, and
  Ranzato}]{lample:unsupervised2:2018}
Guillaume Lample, Myle Ott, Alexis Conneau, Ludovic Denoyer, and Marc'Aurelio
  Ranzato. 2018{\natexlab{b}}.
\newblock Phrase-based \& neural unsupervised machine translation.
\newblock \emph{arXiv}, 1803.05567.

\bibitem[{Lui and Baldwin(2012)}]{lui2012langid}
Marco Lui and Timothy Baldwin. 2012.
\newblock langid. py: An off-the-shelf language identification tool.
\newblock In \emph{Proceedings of the ACL 2012 system demonstrations}, pages
  25--30. Association for Computational Linguistics.

\bibitem[{Luong et~al.(2015)Luong, Pham, and Manning}]{luong:nmt:2015}
Minh-Thang Luong, Hieu Pham, and Christopher~D Manning. 2015.
\newblock {Effective approaches to attention-based neural machine translation}.
\newblock In \emph{Conference on Empirical Methods in Natural Language
  Processing ({EMNLP})}.

\bibitem[{Niu et~al.(2018)Niu, Denkowski, and Carpuat}]{niu:bi:2018}
Xing Niu, Michael Denkowski, and Marine Carpuat. 2018.
\newblock Bi-directional neural machine translation with synthetic parallel
  data.
\newblock \emph{arXiv preprint arXiv:1805.11213}.

\bibitem[{Ott et~al.(2018{\natexlab{a}})Ott, Auli, Grangier, and
  Ranzato}]{ott:uncertainty:2018}
Myle Ott, Michael Auli, David Grangier, and Marc'Aurelio Ranzato.
  2018{\natexlab{a}}.
\newblock Analyzing uncertainty in neural machine translation.
\newblock In \emph{Proceedings of the 35th International Conference on Machine
  Learning}, volume~80, pages 3956--3965.

\bibitem[{Ott et~al.(2018{\natexlab{b}})Ott, Edunov, Grangier, and
  Auli}]{ott:scaling:2018}
Myle Ott, Sergey Edunov, David Grangier, and Michael Auli. 2018{\natexlab{b}}.
\newblock Scaling neural machine translation.
\newblock In \emph{Proceedings of the Third Conference on Machine Translation:
  Research Papers}.

\bibitem[{Papineni et~al.(2002)Papineni, Roukos, Ward, and
  Zhu}]{papineni:bleu:2002}
Kishore Papineni, Salim Roukos, Todd Ward, and Wei-Jing Zhu. 2002.
\newblock {BLEU}: a method for automatic evaluation of machine translation.
\newblock In \emph{Conference of the Association for Computational Linguistics
  ({ACL})}.

\bibitem[{Paulus et~al.(2018)Paulus, Xiong, and Socher}]{paulus:summary:2018}
Romain Paulus, Caiming Xiong, and Richard Socher. 2018.
\newblock A deep reinforced model for abstractive summarization.
\newblock In \emph{International Conference on Learning Representations
  ({ICLR})}.

\bibitem[{Pereyra et~al.(2017)Pereyra, Tucker, Chorowski, Kaiser, and
  Hinton}]{pereyra:regularize:2017}
Gabriel Pereyra, George Tucker, Jan Chorowski, Lukasz Kaiser, and Geoffrey~E.
  Hinton. 2017.
\newblock Regularizing neural networks by penalizing confident output
  distributions.
\newblock In \emph{International Conference on Learning Representations
  ({ICLR}) Workshop}.

\bibitem[{Perez and Wang(2017)}]{perez:augmentation:2017}
Luis Perez and Jason Wang. 2017.
\newblock The effectiveness of data augmentation in image classification using
  deep learning.
\newblock \emph{arxiv}, 1712.04621.

\bibitem[{Poncelas et~al.(2018)Poncelas, Shterionov, Way, de~Buy~Wenniger, and
  Passban}]{poncelas:investigatingBT:2018}
Alberto Poncelas, Dimitar~Sht. Shterionov, Andy Way, Gideon~Maillette
  de~Buy~Wenniger, and Peyman Passban. 2018.
\newblock Investigating backtranslation in neural machine translation.
\newblock \emph{arXiv}, 1804.06189.

\bibitem[{Post(2018)}]{post:sacre:2018}
Matt Post. 2018.
\newblock A call for clarity in reporting bleu scores.
\newblock \emph{arXiv}, 1804.08771.

\bibitem[{Sennrich et~al.(2016{\natexlab{a}})Sennrich, Haddow, and
  Birch}]{sennrich:backtranslation:2016}
Rico Sennrich, Barry Haddow, and Alexandra Birch. 2016{\natexlab{a}}.
\newblock Improving neural machine translation models with monolingual data.
\newblock \emph{Conference of the Association for Computational Linguistics
  ({ACL})}.

\bibitem[{Sennrich et~al.(2016{\natexlab{b}})Sennrich, Haddow, and
  Birch}]{sennrich:bpe:2016}
Rico Sennrich, Barry Haddow, and Alexandra Birch. 2016{\natexlab{b}}.
\newblock Neural machine translation of rare words with subword units.
\newblock In \emph{Conference of the Association for Computational Linguistics
  {(ACL)}}.

\bibitem[{Serban et~al.(2016)Serban, Sordoni, Bengio, Courville, and
  Pineau}]{serban:dialog:2016}
Iulian Serban, Alessandro Sordoni, Yoshua Bengio, Aaron~C. Courville, and
  Joelle Pineau. 2016.
\newblock Building end-to-end dialogue systems using generative hierarchical
  neural network models.
\newblock In \emph{Conference of the Association for the Advancement of
  Artificial Intelligence ({AAAI})}.

\bibitem[{Shaw et~al.(2018)Shaw, Uszkoreit, and Vaswani}]{shaw:relpos:2018}
Peter Shaw, Jakob Uszkoreit, and Ashish Vaswani. 2018.
\newblock Self-attention with relative position representations.
\newblock In \emph{Proc. of NAACL}.

\bibitem[{Sutskever et~al.(2014)Sutskever, Vinyals, and
  Le}]{sutskever:seq2seq:2014}
Ilya Sutskever, Oriol Vinyals, and Quoc~V. Le. 2014.
\newblock Sequence to sequence learning with neural networks.
\newblock In \emph{Conference on Advances in Neural Information Processing
  Systems ({NIPS})}.

\bibitem[{Szegedy et~al.(2015)Szegedy, Vanhoucke, Ioffe, Shlens, and
  Wojna}]{szegedy:inception:2015}
Christian Szegedy, Vincent Vanhoucke, Sergey Ioffe, Jonathon Shlens, and
  Zbigniew Wojna. 2015.
\newblock {Rethinking the Inception Architecture for Computer Vision}.
\newblock \emph{arXiv preprint arXiv:1512.00567}.

\bibitem[{Vaswani et~al.(2017)Vaswani, Shazeer, Parmar, Uszkoreit, Jones,
  Gomez, Kaiser, and Polosukhin}]{vaswani:transformer:2017}
Ashish Vaswani, Noam Shazeer, Niki Parmar, Jakob Uszkoreit, Llion Jones,
  Aidan~N. Gomez, Lukasz Kaiser, and Illia Polosukhin. 2017.
\newblock Attention is all you need.
\newblock In \emph{Conference on Advances in Neural Information Processing
  Systems ({NIPS})}.

\bibitem[{Vincent et~al.(2008)Vincent, Larochelle, Bengio, , and
  Manzagol}]{vincent:2008:icml}
Pascal Vincent, Hugo Larochelle, Yoshua Bengio, , and Pierre-Antoine Manzagol.
  2008.
\newblock Extracting and composing robust features with denoising autoencoders.
\newblock In \emph{International Conference on Machine Learning ({ICML})}.

\bibitem[{Xia et~al.(2017)Xia, Qin, Chen, Bian, Yu, and Liu}]{xia:dual:2017}
Yingce Xia, Tao Qin, Wei Chen, Jiang Bian, Nenghai Yu, and Tie-Yan Liu. 2017.
\newblock Dual supervised learning.
\newblock In \emph{International Conference on Machine Learning ({ICML})}.

\bibitem[{Zhang and Zong(2016)}]{zhang:sourcemono:2016}
Jiajun Zhang and Chengqing Zong. 2016.
\newblock Exploiting source-side monolingual data in neural machine
  translation.
\newblock In \emph{Conference on Empirical Methods in Natural Language
  Processing ({EMNLP})}.

\end{thebibliography}
\bibliographystyle{acl_natbib_nourl}

\end{document}